\documentclass[sigconf]{acmart}
\def\BibTeX{{\rm B\kern-.05em{\sc i\kern-.025em b}\kern-.08emT\kern-.1667em\lower.7ex\hbox{E}\kern-.125emX}}

\usepackage{enumerate}   
\usepackage{wrapfig}    
\usepackage{graphicx}
\usepackage{subfloat}
\usepackage{subfigure}
\usepackage{multirow}
\usepackage{array}
\usepackage{titlesec}
\usepackage{algorithm}
\usepackage{algpseudocode}
\usepackage[compatibility=false]{caption}
\usepackage{algcompatible,amsmath}
\usepackage{caption}  
\usepackage{comment} 
\usepackage{xspace}
\usepackage{fullwidth}
\usepackage{caption}

\titleformat*{\subsubsection}{\large\bfseries}
\begin{document}

\title[]{Deep Geospatial Interpolation Networks}

 \author{ Sumit K. Varshney, Jeetu Kumar, Aditya Tiwari, Rishabh Singh, \\ Venkata M. V. Gunturi, Narayanan C. Krishnan}
\affiliation{%
  \institution{Indian Institute of Technology Ropar, Punjab, India}}
\email{{2019aim1013, 2017csb1083, 2016csb1029, 2016csb1054, gunturi, ckn}@iitrpr.ac.in}






  
\newcommand{\Time}{\textit{T}\xspace}
\newcommand{\Lat}{\textit{M}\xspace}
\newcommand{\Lon}{\textit{N}\xspace}
\newcommand{\Dimension}{\textit{d}\xspace}
\newcommand{\GapSet}{\mathcal{G}\xspace}
\newcommand{\Model}{\textit{DGIN}\xspace}
\newcommand{\Mapfunction}{\textit{M}\xspace}
\newcommand{\ErrorFunction}{\textit{E}\xspace}
\newcommand{\NumTrainSequence}{\textit{L}\xspace}
\newcommand{\InputSequence}{X\xspace}
\newcommand{\GapSequence}{\textit{G}\xspace}
\newcommand{\GapDimension}{k_1 \times k_2 \times \Delta t}
\newcommand{\GapLength}{\Delta t}
\newcommand{\PatchSize}{\textit{p}\xspace}
\newcommand{\encodedRepr}{\textit{E}\xspace}
\newcommand{\latentRepr}{\textit{Z}\xspace}
\newcommand{\sequenceHistory}{\textit{h}\xspace}
\newcommand{\spatialEncoderParameters}{\Phi}
\newcommand{\sequenceModelParameters}{\Psi}
\newcommand{\AttentionModuleParameters}{\mathcal{A}}
\newcommand{\OutputLayerParameters}{\mathcal{O}}
\settopmatter{printacmref=false}
\setcopyright{none}
\renewcommand\footnotetextcopyrightpermission[1]{}
\pagestyle{plain}

\begin{abstract}
Interpolation in Spatio-temporal data has applications in various domains such as climate, transportation, and mining. Spatio-Temporal interpolation is highly challenging due to the complex spatial and temporal relationships. However, traditional techniques such as Kriging suffer from high running time and poor performance on data that exhibit high variance across space and time dimensions. To this end, we propose a novel deep neural network called as \emph{Deep Geospatial Interpolation Network(DGIN)}, which incorporates both spatial and temporal relationships and has significantly lower training time. DGIN consists of three major components: Spatial Encoder to capture the spatial dependencies, Sequential module to incorporate the temporal dynamics, and an Attention block to learn the importance of the temporal neighborhood around the gap. We evaluate DGIN on the MODIS reflectance dataset from two different regions. Our experimental results indicate that DGIN has two advantages: (a) it outperforms alternative approaches (has lower MSE with p-value $< 0.01$) and, (b) it has significantly low execution time than Kriging. 
\end{abstract} 


\maketitle

\vspace{3mm}
\section{Introduction}
\vspace{2mm}
Geo-Spatial interpolation is widely used in various applications such as meteorology, ecology, mining, etc., for filling gaps that arise due to scenarios such as sensor malfunctions and cloud covers. Traditional spatial interpolation methods like Kriging \cite{Kriging1,kriging3}, Inverse Distance Weighting (IDW) \cite{IDW}, Distribution based Distance Weighting (DDW) \cite{DDW}, etc., require high execution time and need to be re-instantiated for every new gap instance. Moreover, our experiments also reveal that Kriging (Ordinary Kriging) has higher error on datasets, which exhibit high variance across space and time dimensions. 

This paper proposes a Deep Geospatial Interpolation Network (DGIN) to interpolate the gaps in the dataset. DGIN is equipped to interpolate based on the availability of past and/or future data surrounding the gap. The DGIN architecture consists of three major components: Spatial Encoder, Sequential Module, and Attention Mechanism. The spatial and sequential modules capture the spatio-temporal properties of the dataset surrounding the gap. The Attention Module determines the importance of the encodings generated by sequence modules for the past and future data surrounding the gap. The weighted encodings are then processed to fill the gap.


The contributions of our work are summarized as follows:
\begin{itemize}
    \item We propose spatial and temporal modules to model the spatial dynamics and non-linear temporal dependencies. 
    \item We propose a generic and computationally efficient approach for interpolation.
    \item We evaluate our model (DGIN) on MOD09A1 data from Australia and Greenland region and observe better results (i.e., lower MSE with p-value $<0.01$) compared with Kriging and STNN \cite{STNN}.
\end{itemize}

\section{Limitations of Related Work}
\vspace{2mm}
Research literature relevant to our work consists of the work done in the areas of traditional spatial statistics \cite{IDW,DDW,Kriging1,kriging3}, spatial data mining \cite{Classification,ijgi4042306}, neural networks for spatio-temporal data \cite{STNN,stnn19}, and computer vision \cite{VI1,VI2}. 

Spatial statistics techniques such as IDW \cite{IDW}, DDW \cite{DDW}, Kriging \cite{Kriging1,kriging3}, and its variants are not suitable for the interpolation problem because of the following reasons: (a) high execution time (in case of Kriging), (b) strong assumptions on the nature of spatial relationships (such as inverse relationship in case of IDW), (c) prior assumption and/or knowledge on statistical properties of data (e.g., precise knowledge of the mean in case of Simple Kriging and presence of constant mean in case Ordinary Kriging). 

Spatial classification techniques \cite{Classification} are not inherently suitable for the interpolation problem as interpolation requires regression.  

Recent works \cite{STNN,stnn19} in the area of neural networks for spatio-temporal data focus on forecasting. STNN \cite{STNN} learns a separate generative model for interpolating every gap. Moreover, our experiments reveal that it performs poorly in case there is high variance (along space and time dimensions) across the gap. Further, STNN is limited to only forecasting. In other words, it cannot take advantage of the future data to fill-in a gap. 

Spatio-temporal interpolation is popularly referred in the computer vision literature as video in-painting \cite{VI1,VI2}. However, these techniques are applicable only on datasets that exhibit very high correlation across space and time dimensions (e.g., videos).

\section{Basic Concepts and Problem Definition}

\begin{definition}{\textbf{Spatio-Temporal Data:}} We are given the input in the form of a spatial-temporal tensor $\InputSequence$ $\in$ $\mathbb{R}^{\Lat \times \Lon \times \Time \times \Dimension}$. Here, \Lat ,\Lon ,\Time, and \Dimension correspond to latitudes, longitudes, the set of discrete time-steps from a time horizon (of interest), and \Dimension corresponds to the number of features measured at each location of our geographic area of interest (over time). 
\end{definition}

\begin{definition}{\textbf{Spatio-Temporal Gap ($\GapSequence_i$):}} A spatio-temporal gap (ST-gap) $\GapSequence_i$ is defined as a $\Dimension$ dimensional hypercube in $\InputSequence$ where all of the $\Dimension$ features are missing continuously for a time interval $\Delta t_i$ (between a specific time $t=\alpha$ and $t=\beta$). An ST-gap is defined to have the dimension $\GapDimension_i$. Note that the values of $\GapDimension$ can vary across different ST-gaps.
\end{definition}

\subsection{Problem Definition}
We are given a set of spatio-temporal tensors, which we refer to as the training set $\mathcal{X}_{tr}=\left\{\InputSequence_1,\ldots,\InputSequence_L\right\}$, containing $L$ ST-gaps. Let the gap in the $i^{th}$ training tensor $\InputSequence_i$ be denoted as $\GapSequence_i$. We are also provided with the feature values $Y_i$ in the ST-gap $\GapSequence_i$ for all the training tensors. 

The overarching problem addressed in the paper is filling the ST-gap $\GapSequence_i$. Specifically, our objective is to learn a function $f$ whose input is a spatio-temporal tensor $X$ and an ST-gap $G$ present in it, and the outputs are the predicted feature values $\hat{Y}$ for the ST-gap.\\
$f:\mathbb{R}^{\Lat \times \Lon \times \Time \times \Dimension}\times \mathbb{R}^{\GapDimension} \longrightarrow \mathbb{R}^{\GapDimension}$. Learning $f$ is formulated as a regression problem minimizing the loss function defined on the $\hat{Y}$ and $Y$, where $f$ is parameterized as a deep neural network. The learned function $f$ is evaluated on an exclusive test set $\mathcal{X}_{te}$. 

\section{Proposed Approach}
\vspace{2mm}
\subsection{Filling an ST-gap}
Consider any arbitrary ST-gap $\GapSequence_i$ of size $\GapDimension$. We re-define $\GapSequence_i$ as $\GapSequence_i =  \bigcup_{j=1}^{k1 \times k2} g_j$. In order words, $\GapSequence_i$ is considered as a collection of ``strands'' of length $\GapLength$. Here, each strand $g_j \in \GapSequence_i$ has a spatial footprint of unit grid location. We fill in $g_j$'s using the data preceding and succeeding the $\GapSequence_i$. Thus, the order in which $g_j$'s are interpolated does not matter. This is an advantage of our model as it makes the training process independent of the size of cuboid $\GapSequence$.

\subsection{Details of Proposed Model}
\begin{figure*}

    \centering
    \includegraphics[width=\textwidth, height=6cm]{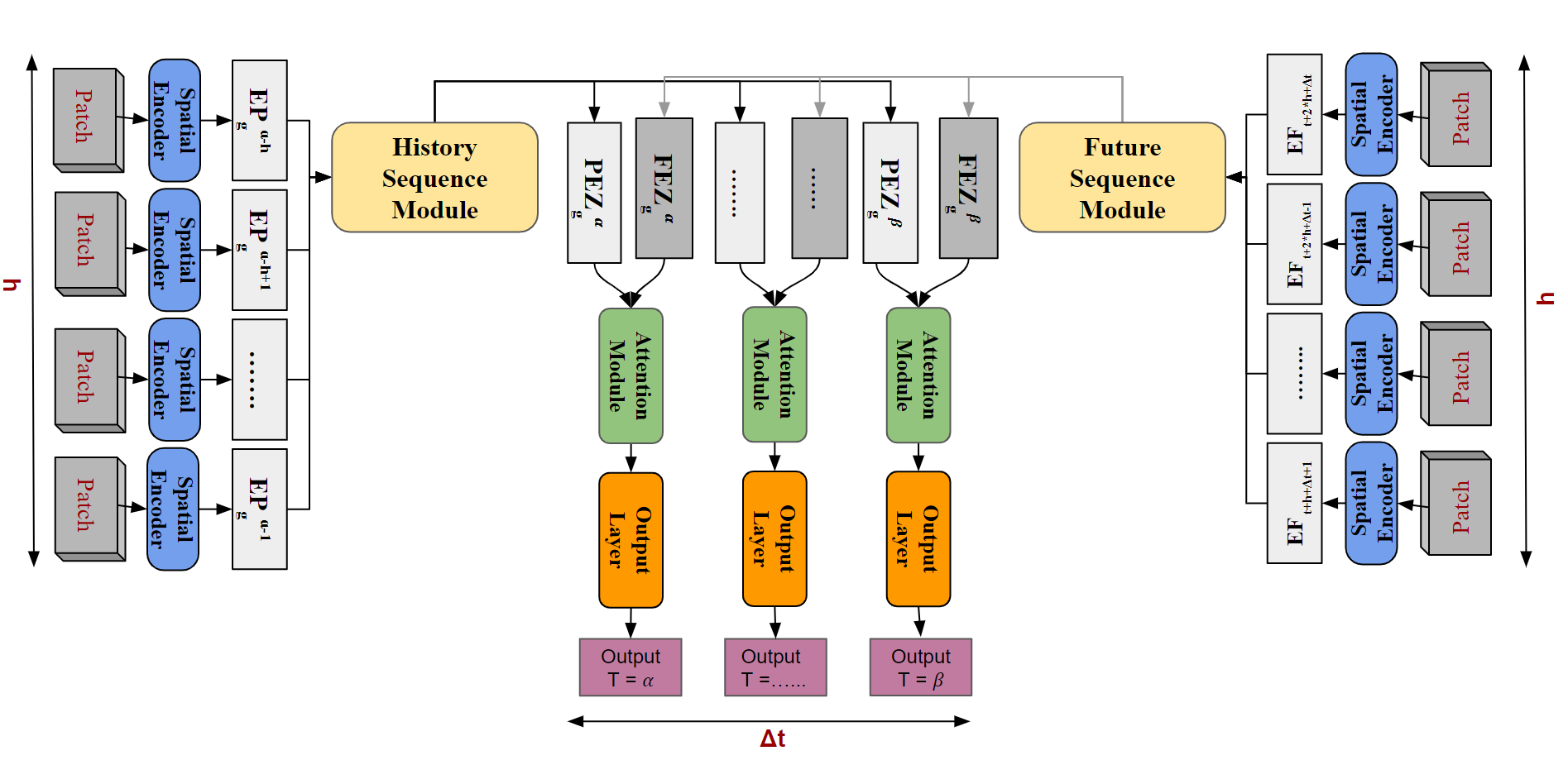}
    \caption{Deep Geospatial Interpolation Network (DGIN) Architecture}
    \label{fig:my_label5}
\end{figure*}

The following are the major components of our model (figure:\ref{fig:my_label5}): (a) spatial encoder, (b) sequence module, and (c) attention module.

\subsubsection{Spatial-Encoder}
\hfill\\
Geospatial data often have spatial dependencies i.e., neighboring locations tend to be more correlated than the far ones. We use this idea to develop our Spatial-encoder. Consider a unit gap $g_j$ ($\in \GapSequence_i$) with a spatial footprint of unit size and a temporal length of $\GapLength$. Now consider a spatial patch $P^t_{g_j}$ (for an arbitrary time point $t$ in $\GapSequence_i$) around the spatial location of $g_j$. The Spatial footprint of patch $P^t_{g_j}$ is a square region (of size $\PatchSize \times \PatchSize$) centered at the spatial location of $g_j$. $\PatchSize$ is a hyperparameter that defines how far spatial relations last. It is reasonable to assume that spatial auto-correlation fades away with distance. This patch ($P^t_{g_j}$) is sent to our spatial encoder module to extract relevant spatial relations and encode the information in the patch and is denoted as $EP^t_{g_j}$ .

Our spatial encoder is a dense layer that compresses spatial patch $P^t_{g_j}$ of size $\PatchSize \times \PatchSize \times \Dimension$ to an encoded patch $EP^t_{g_j}$. Each value of the encoded representation is affected by each value in the input patch. A single layer is used to model all possible relations present inside the patch. Thus, far away relations are also captured directly into the model. While a convolutional layer is also a viable choice, our experiments suggested that a dense layer results in better performance. 

For a given $g_j \in \GapSequence_i$, we would create the following two kinds of spatially encoded patches: (a) \emph{past-patches} and (b) \emph{future-patches}. Past-patches are series of encoded patches $EP^t_{g_j}$s created by encoding the respective $P^t_{g_j}$s for the times in the range $t=\alpha-h$ till $t=\alpha-1$. Similarly, future-patches are series of $EF^t_{g_j}$s are created for the times in the range $t=\beta+1$ till $t=\beta+h$. Here, $h$ is a hyperparameter that denotes the amount of history or future data observed by the model for filling a $g_j \in \GapSequence_i$. And recall that in each $g_j \in \GapSequence_i$, the sensor values were missing (for all features) for the time interval $\GapLength_i = \beta - \alpha$. The sequence module would take multiple $EP^t_{g_j}$s and learn the temporal relationship among them. 
 
\subsubsection{Sequence Module}
\hfill\\
We now describe the process of learning the temporal relationships among the \emph{past-patches} and \emph{future-patches}. 
The sequence module consists of two parts, an encoder and a decoder. The sequence encoder takes in the $h$ number of encoded past-patches ($EP^t_{g_j}$s). Recall that each of these encoded patches feed-in from an instance of the spatial encoder module. The Sequence encoder learns temporal relations among $EP^t_{g_j}$s using Gated Recurrent Units \cite{cho-etal-2014-learning}. 

The output of the sequence encoder is put into a context vector that is used by the sequence decoder. The decoder sequence also uses Gated Recurrent Units (GRU). The decoder GRU ``outputs'' the meta-information (referred to as the forward meta-info) of next $\GapLength_i$ number of patches. These are denoted as $PEZ^t_{g_j}$ (for past sequence model) and $FEZ^t_{g_j}$ (for future sequence model), where $t \in [\alpha, \beta]$. In summary, the sequence module takes in $h$ number of spatially encoded patches from past of $g_j$, learns the temporal relationships among them and, predicts the forward meta-info, $PEZ^t_{g_j}$, of the next $\GapLength_i$ number of patches. A similar process is undertaken for the future-patches of $g_j$. In this case, the sequence module would return backward meta-info $FEZ^t_{g_j}$ for $\GapLength_i$ number of patches. The output from the sequence model is a pair of meta-information ($PEZ^t_{g_j}, FEZ^t_{g_j}$) for each time $t$ in the unit gap $g_j$ of length $\GapLength_i$. These are combined using the Attention Module.

\subsubsection{Attention Module}
\hfill\\
Attention Module is a weighting module to determine the ``importance'' of forward and backward meta-info generated by the sequence decoder to predict all the $\GapLength_i$ number of time coordinates in a unit gap $g_j$ (in a ST-gap $\GapSequence_i$). As one may imagine, the forward meta-info is better suited for predicting the initial part of the gap. Likewise, the backward meta-info would help predicting values towards the end of the gap. 

The attention module takes as input the forward and backward meta-info to first output the attention weights for these vectors. Then, the weighted average of the forward and backward meta-info is computed for every time step. 

Formally, the attention module takes the forward and backward meta-info pair $(PEZ^t_{g_j}, FEZ^t_{g_j})$ at time $t$ and computes score ($e^{P}_{t}, e^{F}_{t}$) by passing it through a dense layer. The dense layer will learn to ``give appropriate'' attention to the forward and backward meta-info vectors relative to $t \in [\alpha, ~\beta]$. We apply softmax activation on ($e^{P}_{t}, e^{F}_{t}$) to produce the attention weights ($w^{P}_{t}, w^{F}_{t}$). We then compute the weighted latent representation ($Z_{g_j}^t$), which is a weighted combination of forward and backward meta-info vectors.

The output of the Attention module at every time step $t$, $Z_{g_j}^t$, is passed through a dense layer to yield the \Dimension-dimensional output.


\subsection{Training the DGIN}
Every module of the DGIN is independently parameterized. Let $\spatialEncoderParameters$ denote the parameters of the spatial encoder ($f_s$), $\sequenceModelParameters$ of Sequence module ($f_t$), $\AttentionModuleParameters$ of attention module ($f_a$) and $\OutputLayerParameters$ of the output layer ($f_o$).
Consider an input pair ${(X_g,g)}$ where $X_g$ is the data including \sequenceHistory patches preceding and succeeding unit gap $g$. Thus, the output of the DGIN can be viewed as a function composition across these modules, i.e., $\hat{g}=f_o(f_a(f_t(f_s(X_g))))$.

The error in the prediction is $Dist(\hat{g},g)$. Given multiple input pairs, training the DGIN involves learning the parameters to reduce the error. Mean Square Error is used as the distance function $Dist$. Let $\mathcal{X}_{tr}$ denote all the training pairs of $(X_g,g)$. The model parses through $\mathcal{X}_{tr}$ in a randomly shuffled order in one epoch. The model is trained for multiple such epochs. We use the ADAM optimizer with a learning rate of 0.0001. The batch size is 128, and the number of epochs is 200. Our model is implemented in PyTorch and is trained on Nvidia GTX 1080 Ti.

\vspace{2mm}

\section{Experiments}
\subsection{Dataset}

The experimental data set is MOD09A1 (\url{https://doi.org/10.5067/MODIS /MOD09A1.006}) (2014-2019). We test DGIN on two regions, Eastern Australia and Greenland. The Australian dataset has a medium variance[$\approx 240$] due to vegetation, whereas the Greenland dataset has a high variance[$\approx 1900$] due to snow cover. Since the feature values in the dataset are too small, values are re-scaled by multiplying with a factor of 100. DGIN is evaluated against Kriging, and STNN \cite{STNN} on three kinds of test gaps: (a) only high variance (across space and time dimensions), (b) only low variance, and (c) mixed (set of both high and low variance gaps). We also vary the spatial footprint of the gaps (50km x 50km, 100km x 100km), which gives us two DGIN (50 and 100) models. DGIN-50 is trained using 28 sequences containing ST-gaps of average dimension $\GapDimension$ = 50km x 50km x 80 days, while DGIN-100 is trained using 28 sequences of average dimension 100km x  100km x 80days. We evaluate the performance of DGIN models using eight sequences containing ST-gaps of spatial footprint (50km x 50km and 100km x 100km) and temporal length of 80 days.

\vspace{-2mm}
\subsection{Comparison with related work}
DGIN was evaluated against the following models:
\begin{itemize}
    \item Mean - A naive approach where for every point inside a ST-gap, we take the nearest non-missing historical points and then take the mean of all the points to fill the ST-gap.
    \item Spatio-temporal Kriging - We are using PyKrige python library for Ordinary 3D Kriging with "variogram\_model" parameter set to "linear". Here, we set time as the third dimension. Due to the high computation time required for the Kriging model, we feed only one (or two) immediate historical information for this model.
    \item STNN - For STNN model, we experimented with 5, 10, 20 and 30 time-steps of past data (with respect to the test gap). However, we report results only for the case of 30 time-steps as their model did not converge for other parameter values i.e., the training error did not decrease with increase in the number of training epochs.  
\end{itemize}

\vspace{-3mm}
\begin{figure*}
  \centering
  \subfigure[Australia (100km x 100km)]{\includegraphics[width=0.45\linewidth,, height=4cm]{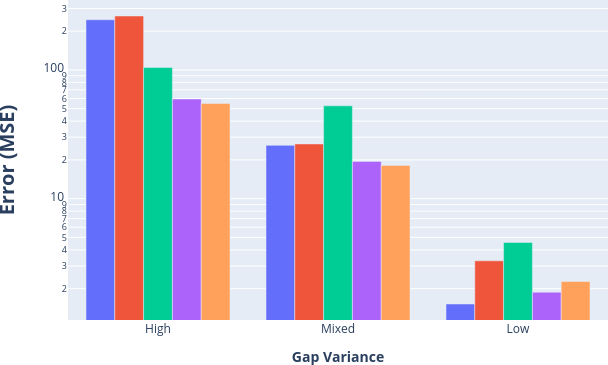}\label{fig:100}}\hspace{5em}%
  \subfigure[Australia (50km x 50km)]{\includegraphics[width=0.45\linewidth, height=4cm]{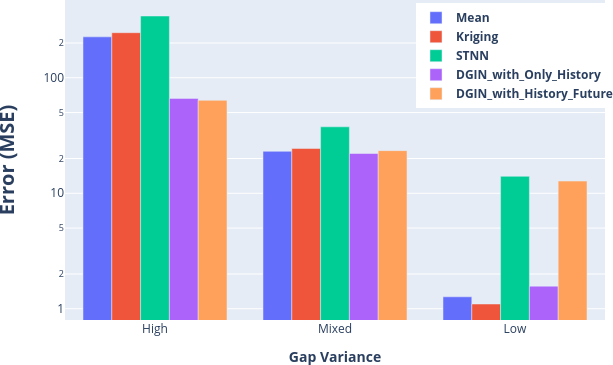}\label{fig:50}}
  \caption{Mean Squared Error(MSE) on different types of ST-gaps over the Australia region. }
  \label{fig:dataset_region}
\end{figure*}
 
\begin{figure*}
  \centering
  \subfigure[Greenland (100km x 100km)]{\includegraphics[width=0.45\linewidth, height=4cm]{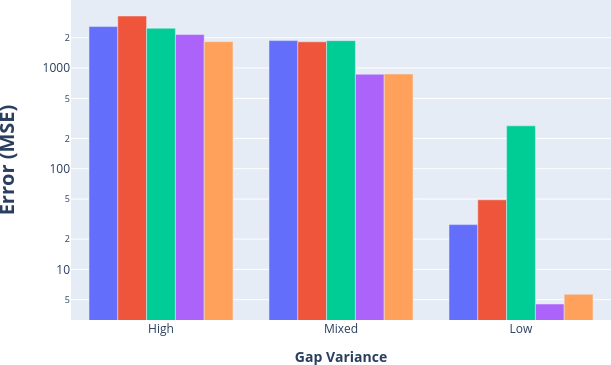}\label{fig:G100}}\hspace{5em}%
  \subfigure[Greenland (50km x 50km)]{\includegraphics[width=0.45\linewidth, height=4cm]{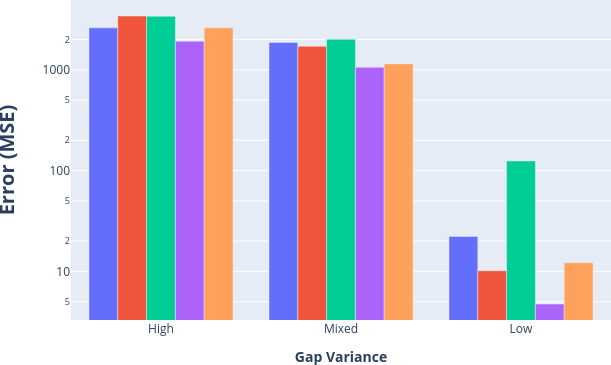} \label{fig:G50}}
  \caption{Mean Squared Error(MSE) on different types of ST-gaps over the Greenland region.}
  \label{fig:dataset_region1}
\end{figure*}


\vspace{-2mm}
\subsection{Results and Observations}
\begin{itemize}
    \item DGIN outperforms Kriging and STNN model in all types of ST-gaps indicating that it can learn complex spatio-temporal patterns present in the ST-gaps (figure-\ref{fig:dataset_region},\ref{fig:dataset_region1}). 
    \item We used $t$-Test to test the significance of DGIN compared to Kriging and STNN. The p-value is less than 0.001 and 0.01 for kriging and STNN respectively, demonstrating that DGIN is statistically outperforming Kriging and STNN.
    \item If historical and future information do not correlate with the ST-gap, then DGIN does not perform well compared to STNN.
    \item The average computation time taken by Kriging is $\approx 1000$ minutes, and STNN is $\approx 35$ minutes for filling a single ST-gap ($\GapSequence_i$). In comparison, DGIN takes $\approx 25$ minutes to train on a dataset containing 28 training ST-gaps. Once the model is learned, it can be used to make multiple inferences in few seconds.
\end{itemize} 

\vspace{-1mm}
\begin{table}
\centering

\begin{tabular}{|l|l|} 
\hline
Method                      & Computation Time(in secs)  \\ 
\hline
Mean                        & 10                         \\ 
\hline
Ordinary Kriging            & 60147                      \\ 
\hline
STNN                        & 2192                       \\ 
\hline
DGIN\_With\_only History    & 1226                       \\ 
\hline
DGIN\_With\_History\_Future & 1548                       \\
\hline
\end{tabular}
\caption{Average Computation Time}
\vspace{-6mm}
\end{table}

\vspace{-1mm}
\section{Conclusion}
We proposed a Deep Geospatial Interpolation Network (DGIN) to interpolate the given dataset's missing information in the ST-gaps. DGIN model has spatial and temporal modules with an attention mechanism to model the complex spatio-temporal relationships present in the data. Experiments on two different region datasets explain that DGIN achieves better results than the alternative approaches. Furthermore, the average computation time for training DGIN is much less than the Kriging and STNN. In the future, we will incorporate the periodicity present in the dataset for better interpolation of information.



\vspace{-1mm}


\balance{}

\bibliographystyle{ACM-Reference-Format}
\bibliography{main} 

\end{document}